\documentclass[twoside,12pt,english,nontitlepage,openany]{article}
\usepackage{babel}
\usepackage[utf8]{inputenc}
\usepackage{color}
\definecolor{marron}{RGB}{60,30,10}
\definecolor{darkblue}{RGB}{0,0,80}
\definecolor{lightblue}{RGB}{80,80,80}
\definecolor{darkgreen}{RGB}{0,80,0}
\definecolor{darkgray}{RGB}{0,80,0}
\definecolor{darkred}{RGB}{80,0,0}
\definecolor{shadecolor}{rgb}{0.97,0.97,0.97}
\usepackage{graphicx}
\usepackage{wallpaper}
\usepackage{wrapfig,booktabs}

\usepackage{fancyhdr}
\usepackage{lettrine}
\input Acorn.fd
\newcommand*\initfamily{\usefont{U}{Acorn}{xl}{n}}

\usepackage{geometry}
\usepackage[
final,
stretch=10,
protrusion=true,
tracking=true,
spacing=on,
kerning=on,
expansion=true]{microtype}

\usepackage{fourier-orns}

\newcommand{\ornamento}{\vspace{2em}\noindent \textcolor{darkgray}{\hrulefill~ \raisebox{-2.5pt}[10pt][10pt]{\leafright \decofourleft \decothreeleft  \aldineright \decotwo \floweroneleft \decoone   \floweroneright \decotwo \aldineleft\decothreeright \decofourright \leafleft} ~  \hrulefill \\ \vspace{2em}}}
\newcommand{\ornpar}{\noindent \textcolor{darkgray}{ \raisebox{-1.9pt}[10pt][10pt]{\leafright} \hrulefill \raisebox{-1.9pt}[10pt][10pt]{\leafright \decofourleft \decothreeleft  \aldineright \decotwo \floweroneleft \decoone}}}
\newcommand{\ornimpar}{\textcolor{darkgray}{\raisebox{-1.9pt}[10pt][10pt]{\decoone \floweroneright \decotwo \aldineleft \decothreeright \decofourright \leafleft} \hrulefill \raisebox{-1.9pt}[10pt][10pt]{\leafleft}}}



\newcommand{\estcab}[1]{\itshape\textcolor{marron}{\nouppercase #1}}

\usepackage{lipsum}
\usepackage[hang,splitrule]{footmisc}
\usepackage{biblatex}
\usepackage{chngcntr}

\bibliography{sources}

\title{Questions to Guide the Future of Artificial Intelligence Research}
\author{Jordan Ott}
\date{}

\begin{document}
\maketitle
\ornamento
\lettrine[lines=3]{\initfamily\textcolor{darkgreen}{T}}{he field of machine learning} 
has focused, primarily, on discretized sub-problems (i.e. vision, speech, natural language) of intelligence. Neuroscience tends to be observation heavy, providing few guiding theories. It is unlikely that artificial intelligence will emerge through only one of these disciplines. Instead, it is likely to be some amalgamation of their algorithmic and observational findings. As a result, there are a number of problems that should be addressed in order to select the beneficial aspects of both fields. In this article, we propose leading questions to guide the future of artificial intelligence research. There are clear computational principles on which the brain operates. The problem is finding these computational needles in a haystack of biological complexity. Biology has clear constraints but by not using it as a guide we are constraining ourselves. 

\ornamento

\section*{Key Points}
\begin{enumerate}
    \item Restrictions of gradient descent limit the space of possible models
    \item Computational principles can be derived from the brain without following every convoluted detail
    \item The locality of current paradigms opposes evidence from the brain 
    \item Short comings of Classical AI systems are still present in Deep Learning
    \item Parts must be understood in the context of the whole not in isolation
    \item Complex internal dynamics give rise to abstract behaviors 
    \item Approximating a behavior is not the same as the behavior itself
\end{enumerate}

\newpage

\section{Introduction}

\lettrine[lines=3]{\initfamily\textcolor{darkgreen}{T}}{he goal} 
of artificial intelligence (AI) has always been to create artificial agents (machines, computers, robots), endowed with \textit{intelligent} capabilities \cite{turing2009computing}. Discussions on \textit{what} intelligence \textit{is} have yielded little substantive guidelines. This absence leaves researchers without a goal to work towards. As a result, researchers began to discretize intelligence. The field of machine learning has been composed of sub-problems,  identified to appease the foggy definition of intelligence. Researchers first identify a capability of human intelligence and segment this area to tackle the problem independently. ``Humans perceive visual stimuli in their environment'' $\rightarrow$ computer vision \cite{ott2018deep, ott2018learning, lecun2015deep, krizhevsky2012imagenet}. ``Humans use language to communicate abstract ideas" $\rightarrow$ natural language processing \cite{bahdanau2014neural, collobert2008unified}. ``Humans make sense of various frequencies of air vibrations" $\rightarrow$ speech recognition \cite{hinton2012deep}. All fields in machine learning were derived by this process. 

We now have a discrete number of sub-fields within machine learning. Each of these sub-disciplines has its own specialized approach with curated datasets and benchmark models. This has led to accuracy competitions which entrenches competitors - researchers - to out perform one another. Hundreds of hours spent fighting over an additional 0.5\% on the test set accuracy \cite{russakovsky2015imagenet, zhongSeq2SQL2017}. These competitive rankings are beneficial to the engineering progress. Yielding deeper networks and more efficient solutions at inference time. However, one can argue that it has not produced a better understanding of intelligence or brought us closer to unifying these sub-disciplines.  

Within the field of neuroscience the same discretization is taking place. Cortices are segregated and studied independently. Neuroscience produces enormous amounts of detailed observational data. However, it lacks unifying theories explaining experimental results. Of particular interest to this article is the work being done to unify the fields of machine learning and neuroscience. Specifically topics like spiking neural networks \cite{diehl2015unsupervised}, biologically plausible backpropagation \cite{lillicrap2016random, baldi2018recirc}, bio-inspired architectures \cite{ott2019learning, bengio2015towards}, and treating neurons as individual actors pursuing their own self-interest \cite{ott2019neuronrl, ott2020hayek}. These solutions offer algorithmic theories grounded in biologically accurate implementations. Further, there have been numerous calls to integrate neuroscience into machine learning \cite{hassabis2017neuroscience, marblestone2016toward}. Demand for this integration suggests that machine learning alone cannot offer a complete solution and neuro-scientific principles may remedy some of deep learning's pitfalls \cite{marcus2018deep}.

It is unlikely that artificial intelligence will emerge through only neuroscience or machine learning. Instead, it is likely to be some amalgamation of their algorithmic and observational findings. In the following sections we identify major questions which aim to guide the future of artificial intelligence research. We will focus primarily on the construction of deep artificial neural networks with the exception of Section \ref{symbolic}, which provides a discussion of Symbolic AI and hybrid approaches. These questions cover machine learning, neuroscience, and the philosophy surrounding their intersection. Others have put forth guiding principles to build artificial agents \cite{lake2017building, marblestone2016toward}. However, these papers often speak in vague generalities that are likely to miss implementation level details required to accomplish this enormous goal. The purpose of this paper is to not only ask broad questions but also to address specific details we believe to be essential elements in the creation of artificial agents.

\section{Can intelligence be discretized?}
\label{descritize}
\lettrine[lines=3]{\initfamily\textcolor{darkgreen}{T}}{he discretization} 
of intelligence, as previously mentioned, refers to breaking human capabilities into discrete sub-problems and tackling them independently. This is evident by the phrasing, ``humans can do $X$, we can build a machine that does $X$". This process is exemplified by the recent work in reinforcement learning, ``Humans can ...":

\begin{itemize}  
    \item learn from past experiences $\rightarrow$ Experience Replay \cite{lin1991programming}
    \item learn from past mistakes $\rightarrow$ Hindsight Experience Replay \cite{andrychowicz2017hindsight}
    \item learn from their larger mistakes $\rightarrow$ Prioritized Experience Replay \cite{schaul2015prioritized}
    \item learn hierarchical tasks $\rightarrow$ Hierarchical Reinforcement Learning \cite{singh1992reinforcement}
    \item generate goals intrinsically $\rightarrow$ Intrinsic motivation \cite{barto2013intrinsic}
\end{itemize}

Following this process we see that nearly all fields - search, planning, vision, speech recognition, natural language processing - were derived in this manner. Each of these models aim to solve an increasingly specific niche of human cognition. Naturally, this is the way to approach problems as broad and complicated as \textit{intelligence}. However, it is important to question whether this approach will lead to human level cognition, or solely the domain specific solutions we have currently (coined \textit{narrow AI} \cite{goertzel2007artificial}). The danger in this approach is attempting to laboriously list all human capabilities, formulate explicit models for each faculty (i.e. direct cost functions, see Section \ref{direct_vs_indirect}), and approximate their functionality (Section \ref{anthro_and_approx}) without regard for their interaction in the system at large.

Intelligence doesn't arise by solving perception or pattern recognition problems such as vision or audition. We need only look to our companions across the animal kingdom for evidence. Human intelligence appears special, at a cognitive-functional level, because of our ability to integrate information over time within a cognitive hierarchy. Meaning, not only can we solve these perception and motor tasks but we have cognitive layers proceeding that. Layers (not physical layers but abstract processing layers) allowing attention, memory, and introspection - so that one may attend to his own thoughts or even thoughts about thoughts. 

In this way models that solve only a single sensory modality (e.g. vision) may critically miss the point. This process will yield well engineered solutions for that specific domain but we must not allow ourselves to extrapolate its properties beyond that (see Section \ref{anthro_and_approx} for a description of anthropomorphism). Making better vision systems that increase test set accuracy are somewhat orthogonal to intelligence. Functional properties should not be cut off and implemented in isolated ways. Instead, it is important to understand how these systems work in the context of the whole (Section \ref{smi}). 

To move beyond domain specific solutions we must incorporate mechanisms like attention \cite{xu2015show, luong2015effective}, association \cite{santoro2017simple, santoro2018relational}, and memory \cite{graves2014neural, sukhbaatar2015end}. Currently, methods like deep learning attention and association still work within an isolated domain. These higher cognitive layers should be multi-modal and introspective. The catch for deep learning is that the more abstract the model, the less clear the cost function becomes (Section \ref{cost}).

Machine learning, to this point in time, relies on discretizing intelligence. It is not clear whether this will lead to ultimate success. At this time it yields narrow AI - domain specific solutions. The ultimate point is that the human brain is able to search, plan, see, hear, learn from mistakes, attend to memories, and complete an endless list of other cognitive tasks without explicit models. The brain solves search without a tree, plans without a directed graph, and learns without gradient descent. Trees, graphs, and optimization are mathematical formalisms and human abstractions to describe specific problems. The current thinking is that we will master each of these problems independently then stitch the resulting models together. Again, critically missing the point that the brain solves all cognitive tasks with the same basic circuitry \cite{mountcastle1957modality}. That discretizing intelligent behaviors is not synonymous with the system as a whole. 

\section{Why are we so dense?}
\label{encode}

\begin{figure}
    \centering
    \includegraphics[width=\linewidth]{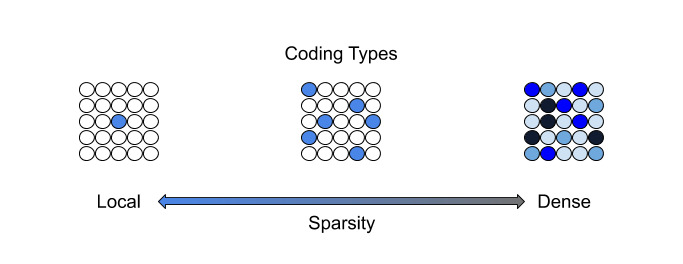}
    \caption{\footnotesize Coding schemes can be defined on a spectrum of sparsity. At one extreme, complete sparsity, are local codes. Where a single node is active. At the other extreme, no sparsity, are dense codes. Here all nodes are active at all times. In between these two are the spectrum of sparse codes defined by the percentage of active nodes in the population.  }
    \label{fig:sparsity_spectrum}
\end{figure}

\lettrine[lines=3]{\initfamily\textcolor{darkgreen}{C}}{oding schemes} for neural networks form a spectrum, defined by the sparsity of their nodes (Figure \ref{fig:sparsity_spectrum}). Tradeoffs in learning speed, storage capacity, and fault tolerance are made as one moves across this spectrum. 

The sparsest type of code is the local code. This is achieved by activating a single node out of $n$ in the population. Encoding information via this process yields a capacity of $n$. The sacrifice of capacity is remedied by the speed of learning and the high degree of fault tolerance \cite{foldiak2003sparse}. On the other end of the spectrum, with no sparsity, are dense codes. In this coding scheme all nodes are active at all times, encoding information in real numbers. While dense codes can yield near infinite capacity they are hindered by slow learning speeds and susceptibility to noise \cite{foldiak2003sparse}. In between local and dense codes are the spectrum of sparse codes, where the degree of sparsity is imposed by the percentage, $p$, of active nodes in a population. The  capacity of a sparse code is given by, $ {n \choose n*p}$. Again, there is a tradeoff between capacity, learning speed, and fault tolerance. As $p$ increases towards 50\% (the maximum of $ {n \choose n*p}$) the capacity increases, however, the speed of learning decreases. The process of evolution may well have found the optimal solution to this tradeoff. Producing a neocortex with a high degree of sparsity resulting in fast learning speeds while still retaining high storage capacity.

From neuroscience's point of view, spikes arose as an evolutionary solution to information transmission at a distance. Spikes are essential to transmit a signal from a neuron to a distant neuron. Graded potentials of ions are too weak and diffuse too rapidly to carry information over such distances \cite{wolf2004neuroscience}. 

From machine learning's perspective, spikes are a biological constraint imposed by the resources available to organic matter. As spikes or binary thresholds are not differentiable, artificial neural networks rely predominantly on dense coding. However, a growing body of work indicates that spiking networks or increased sparsity may provide robust capabilities \cite{ahmadcan, wu2018spiking}. This desired trait is derived from the properties of sparse codes \cite{foldiak2003sparse}, and are beneficial for memory capacity \cite{brunel2004optimal, foldiak2003sparse} and learning speeds \cite{schweighofer2001unsupervised}. Dense codes may be hindered by their slow learning \cite{foldiak2003sparse} and need to update all representations with every backward pass \cite{goodfellow2013empirical,kirkpatrick2017overcoming}.

The coding scheme is one of the most fundamental aspects of artificial networks. Deep learning has mostly avoided spiking and threshold models because of their non differentiability. Meaning, they can't be trained via backpropagation. However, as gradient descent is unlikely to be implemented in biological circuits, exploration of other learning rules (Section \ref{learning_rules}) that take advantage of sparse codes may provide beneficial advancements.

\section{Can you turn the noise down?}
\label{robust}
\lettrine[lines=3]{\initfamily\textcolor{darkgreen}{T}}{he sensitivity} of models to noise and adversarial attacks has gained tremendous interest of late \cite{marcus2018deep}. Creating robust models is of the upmost importance, as increasing responsibilities are granted to these models - autonomous vehicles, medical, security. In this section we review the brittleness of deep learning models, highlight capabilities of the mammalian cortex with regard to invariant transformations, and provide examples from neuroscience on potential solutions.

\subsection{Adversarial Attacks}
\label{adversarial}
A growing field in deep learning is adversarial attacks. Artificial networks are shown to be extremely fragile to perturbations in the input space \cite{yuan2019adversarial, akhtar2018threat}. In extreme cases, changing one pixel results in a catastrophic change in the output of the network \cite{su2019one}. These attacks have obvious consequences in domains such as self driving cars, medical diagnosis, and security. This susceptibility leads one to question how the brain can be so robust. 

Ignoring cognitive development periods and neurological disorders, the brain does not fail to learn. Once it has, catastrophic forgetting \cite{goodfellow2013empirical, kirkpatrick2017overcoming} is not an issue (for many decades at least) and it is not susceptible to adversarial attacks. How does the brain achieve such feats? 

Section \ref{encode} detailed examples of the robust properties of sparse codes. Sparse representations, like those in spiking neural networks, are extremely robust to noise \cite{ahmadcan, wu2018spiking}.  Conversely, in machine learning a neural network - with dense coding - is trained, in a supervised manner, to approximate some function. More explicitly, the weights are modified by a gradient descent algorithm to minimize an objective function over all points in the training set. It is easy to see from adversarial examples, approximating a behavior may miss critical aspects of what the behavior encompasses (Section \ref{anthro_and_approx}). As soon as an input deviates from the training data manifold, even by a single pixel \cite{su2019one}, chaos can ensue. 

\subsection{Invariant Representations}
\label{invariance}

The brain is invariant to transformations of many kinds - all without \textit{explicit} models to do so. Mathematically, an invariant function, $f$, satisfies Equation \ref{eq:invariance}. In which $f$ produces the same result on an input, $X$, as it does under some transformation, $T(X)$. Described below are a variety of transformations that $T$ may take in the physical world.    

\begin{equation}
\label{eq:invariance}
    f(T(X)) = f(X)
\end{equation}

\subsubsection*{Example of Invariance}

Input to the brain may be highly corrupted from varying degrees of noise in the environment. For example, while riding a train your friends voice is dampened by the screech of the wheels, the outside wind, and other passenger's conversations. However, your conversation carries on without a hitch. Similarly, visual information may appear at a range of qualities. A low quality, pixelated, video is still interpretable to viewers. Driving in stormy conditions results in highly corrupted visual information. However, humans can navigate and identify objects visually regardless of the elements. This invariance to \textbf{noise} is desirable trait currently lacked by artificial systems.

\subsubsection*{Other Transformations}
\textbf{Temporal:} Auditory and visual stimuli make take place at a variety of speeds - hearing a speech at 2X speed or watching a video in slow motion. \textbf{Size}: Objects may be small, large or anywhere in between - a toy house in a doll set versus a house tens of meters high. \textbf{Shape}: Physical objects often undergo distortions - kittens curled in a ball instead of lying flat. \textbf{Translation:} Entities can appear at various locations in space - a chair in the left of your visual field is the same when it appears on your right. \textbf{Rotation:} Many objects only appear in a single orientation, yet on rare occurrences undergo rotations - a vehicle in a bad accident may flip upside down. Despite the possible transformations that entities may undergo the brain is robust to all of them.

\subsubsection*{Deep Learning Solutions to Invariance}
Deep learning attempts to solve invariance - \textit{external} to the model - via manipulations to the data. Data augmentation involves performing a variety of perturbations and contortions to the input space in order to increase the effective amount of training examples, thereby decreasing variance in model performance. This process is how multiple types of invariance are addressed. For example, artificial networks are not natively size invariant.\footnote{Some size invariance is achieved via pooling operations, however, it is very limited} In order to combat this deficiency, images are augmented by random cropping - zooming in and out by varying degrees. Size augmentation provides networks with training examples at multiple scales thereby remedying this invariance. 

This approach to invariance is an attempt to approximate Equation \ref{eq:invariance} (Section \ref{anthro_and_approx}). Deep learning's solution results in manipulations to the data - external to the model. This is in contrast to biological mechanisms which aid in robust invariance, that take place within the model. 

\subsection{Robust Properties from Biology}
\label{robust_biology}
Examples of homeostatic properties can be gleamed from neuroscience. When activated, inhibitory neurons cause a decrease in activity of their post-synaptic contacts. As a result, when excitatory neurons become active they can trigger inhibitory interneurons, thereby reducing the activity of neighboring neurons. This ensures that a select few neurons will code a particular stimuli (i.e. sparse codes). Additionally, dynamic thresholds allow populations of neurons to maintain a constant sparsity level \cite{bienenstock1982theory}. As activity increases, the threshold rises - requiring more stimuli to trigger an action potential. As activity decreases, the threshold decreases.

Following these examples, small world networks provide an economic balance between segregation and integration \cite{lord2017understanding}. Synapto genesis and pruning constantly modify the physical wiring within the brain. No connection, or series of synapses, is guaranteed to remain static for long periods of time. As a result, many pathways are formed to provide redundancy, amplifying robustness of the overall network. Synaptic weights regularize themselves during the learning process. When synaptic weights are updated, the ratio of weights are preserved \cite{turrigiano1999homeostatic}. This yields weight normalization and a means to control exploding or vanishing activities.

These mechanisms control the amount of neuronal activity which constrains resources, stops run away excitation, and enforces sparse codes. These properties serve to stabilize cortical networks which specifically aids in noise robustness. Networks with internal regularization mechanisms are able to self correct in order to maintain bounded network behavior. These checks on activity guarantee dynamics even in the presence of adversarial or noisy stimuli.

\section{Where's my reward?}
\label{rewards}

\lettrine[lines=3]{\initfamily\textcolor{darkgreen}{T}}{his section} involves the locality of rewards and how externality affects learning. This topic also envelops preference learning, intrinsic motivation, and reinforcement learning. 

First, a distinction should be made between two types of rewards. There are low level rewards given via processes in the autonomic system - we have relatively little control over their effect. Things such as the taste of food, refreshment of water, and the pain of touching a hot pan fall under this category. High level rewards are conceptualized abstractions from the purely physical reality. For example, there is nothing in our anatomy to suggest a sphere crossing a line should elicit euphoria. However, millions of individuals and entire countries feel tremendous reward, or lack there of, when this happens in a soccer match. These rewards are manufactured entirely within the agent. It is these high level rewards we will focus on, as we have control in shaping them and their ultimate effect on us.

Biologically, rewards are generated \textit{internally} - within the organism. The amount of positive or negative effect this has on the organism is subjective and depends entirely on how the organism's nervous system processes the stimuli. The reward system in mammals can be traced to the ventral tegmental area (VTA) and a group of neurons that produce the neurotransmitter dopamine \cite{wise1998drug, wolf2004neuroscience}. Neurons in VTA make projections to the nucleus accumbens (via the mesolimbic pathway) and to the cerebral cortex (via the mesocortical pathway) \cite{wise1998drug}. The latter, plays a role in neocortical functions like learning and planning. 

The purpose of dopamine is highly contested. It is often thought dopamine is a neuromodulator, involved in multiple aspects of cognition (learning, pleasure, reward) \cite{berridge2007debate}. It has been found that dopamine neurons encode reward prediction errors in reference to expected and unexpected stimuli \cite{ljungberg1992responses, schultz1993responses}.  As a result, dopamine is often acclaimed by reinforcement learning for how it fits with theories like temporal difference learning \cite{sutton2018reinforcement}.

In machine learning, goals are determined by the learning paradigm - supervised, unsupervised, and reinforcement learning. In this sense goals are explicit and fixed throughout training and testing. Where the goal is determined by \textit{directly} optimizing a cost function (Section \ref{direct_vs_indirect}). Rewards in reinforcement learning are generated \textit{externally} and are obtained through interaction with the environment. For example, a reward given by the environment could be the points scored in a video game or a binary win/loss in the case of chess. The agent's goal is, typically, to maximize the expected sum of future rewards \cite{sutton2018reinforcement}. Intrinsic motivation deals with endowing agents with the ability to set their own goals \cite{chentanez2005intrinsically}. This feature represents an important shift towards \textit{internally} specified rewards.

Internally generated rewards, in the case of biology and intrinsic motivation, allow representations, of how beneficial a stimulus is, to be altered through the course of learning. So that feedback from other cortical areas is likely to have an impact on the future release of dopamine in response to the same stimulus \cite{carr2000projections}. In this way, rewards from dopamine help shape learning in other parts of the brain and those parts, in turn, help shape the release of dopamine. This feedback process is something artificial networks currently lack (Section \ref{feedback}).

We can, and should, provide agents with external stimuli, however, interpretation and dissemination of that stimuli should occur internally. This paradigm stands diametrically opposed to deep reinforcement learning, where agents are trained to explicitly maximize predefined, external, rewards. In order to create intelligent agents with autonomy we must move away from externally specified goals. Such as ones prescribed by the training paradigm (i.e. supervised learning and reinforcement learning), that lead to narrow AI. External reward paradigms force the agent to learn a mapping from input stimulus directly to the external reward. This may reduce an agent's ability to generate causal models, as the value of a stimulus is predetermined and immutable by the agent. Therefore, agents must be designed with the ability to asses stimuli and assign arbitrary \textit{rewards} to it. In this way, agents can internally construct goals and take actions in order to satisfy them.

\section{What form should learning rules take?}
\label{learning_rules}
\lettrine[lines=3]{\initfamily\textcolor{darkgreen}{A}}{ learning rule} (like Hebbian \cite{hebb1949organization}, Oja \cite{oja1982simplified}, Spike Timing Dependent Plasticity \cite{markram1995action, gerstner1996neuronal}, etc.) specifies the dynamics for synaptic weight modifications in networks of connected neurons. Learning rules form a two dimensional manifold, in which they vary based on locality of information in \textit{space} and \textit{time} \cite{baldi2016theory}. Discussion in this area leads to two, closely entangled, questions. First, are deep targets\footnote{The desired output of a network given a specific input. For example, in the supervised learning paradigm this trains a network to approximate the function mapping $f:X \rightarrow Y$} necessary for learning? Second, is a form of credit assignment\footnote{A way of determining a synaptic weights contribution to the output of a network.} necessary for learning in deep networks? 

\subsection{Are deep targets necessary for learning?}
\label{deep_targets}
The efficacy of deep targets in conjunction with the backpropagation algorithm is well documented \cite{lecun1998gradient, krizhevsky2012imagenet}. However, backpropagation's biologically implausible nature has been questioned since its inception \cite{stork1989backpropagation,liao2016important, crick1989recent}, calling into question its superiority as a learning algorithm. Many models have been proposed to alleviate these implausible requirements \cite{baldi2018recirc, lillicrap2016random, bengio2015towards}. Despite efforts made to reduce backpropagation's implausibilities, the problem of deep targets still looms as a necessity. 

From a computational-biological standpoint, deep targets are possible, however, it is unclear where the target signal would originate from. It has been hypothesized a signal could be derived from motor commands, reward prediction errors \cite{glimcher2011understanding, sutton2018reinforcement}, input stimulus reconstruction \cite{bengio2015towards, baldi2018recirc}, or local losses \cite{nokland2019training, jaderberg2017decoupled}. However, only reward prediction errors have been confirmed experimentally \cite{ljungberg1992responses, schultz1993responses}. But there is not sufficient evidence to claim it serves as a deep target for cortical networks. 

It has been shown that learning rules, local in space and time (i.e. not backpropagation), are insufficient to learn deep targets \cite{baldi2016theory}. Recently, unsupervised models trained with local learning rules (local in space and time) have shown promising performance, comparable with models utilizing deep targets \cite{diehl2015unsupervised}. This result leads one to question the necessity of deep targets. While local learning rules cannot explicitly minimize a cost function they may still approximate it indirectly (Section \ref{direct_vs_indirect}). 

Advancement in this area could lead to a number of innovations, most readily in the data space. Currently, in deep learning curating large labeled datasets provides a tremendous overhead cost to model construction \cite{russakovsky2015imagenet, ott2018learning, ott2018deep}. Thus, the realization of algorithms that learn via completely unsupervised-local learning rules could provide extreme value to machine learning. 

\subsection{Is a form of credit assignment necessary for learning in deep networks?}
\label{credit_assignment}
If deep targets are not necessary one may ask if credit assignment is even necessary. We must be specific in regards to defining credit assignment and the locality at which it operates (i.e. where it falls on the spectrum of \textit{space} and \textit{time}). 

In artificial neural networks, credit is typically assigned via backpropagation \cite{werbos1974beyond, rumelhart1985learning} in a computation graph. Where error values of the deep targets are computed with respect to the weights, via the chain rule. As information is propagated backwards through the graph (neural circuitry), it becomes locally available, in space, to neurons. However, for layers to compute gradients requires a transpose of the weight matrix - non-local spatial information, known as the weight transport problem \cite{grossberg1987competitive}. Additionally, non-local temporal information from the forward pass is required to compute derivatives and update weights.\footnote{By the time error signals are propagated backwards through the network, new feedforward input may have arrived at the current layer. Making it impossible to compute derivatives with respect to the original input.} These restrictions make backpropagation non-local in space and time. Alleviating the weight transport problem leads to algorithms which are local in space but not in time \cite{lillicrap2016random}.

More biologically constrained networks, implement rules such as Hebbian learning \cite{hebb1949organization} or Spike Timing Dependent Plasticity (STDP) \cite{markram1995action, gerstner1996neuronal} which do not require deep targets for learning.  These learning rules only depend on the input from the pre-synaptic cell and output from the post-synaptic cell. These properties make them completely local in both space and time. 

It appears highly unlikely that neural mechanisms could learn complex feature representations without a form of credit assignment. All neural algorithms\footnote{The one exception to credit assignment in artificial networks is found in evolutionary algorithms \cite{ilonen2003differential, salimans2017evolution}. Here populations of neurons are evolved based on their \textit{fitness} to the environment. Each population undergoes slight mutations to their weights and the most advantageous models are selected for the next iteration. However, while this process of evolution can lead to good performing models it does not allow the network to continually learn after evolution. It appears this property of continual learning still requires a form of credit assignment.} to date involve updating a synaptic weight in accordance to some metric (global in the case of backpropagation and local for Hebbian and STDP) \cite{oja1982simplified, werbos1974beyond, rumelhart1985learning, gerstner1996neuronal, markram1995action, hebb1949organization}. This can be seen in the most naive learning algorithm - random weight perturbations. In this algorithm, a single weight is randomly modified, the output of the network observed, and then modified once more to tune the output accordingly. In this simple case, credit is assigned to the individual weight on how a small change to its value effects the output of the network. While this algorithm is naive and exponentially slow (backpropagation is more efficient than random perturbations by a factor of the number of connections \cite{hinton2007recognize}), it nevertheless displays credit assignment.

Given the efficiency at which the brain learns and stores information, there must be a type of synaptic credit assignment taking place. The question for future research is: how is this credit assigned and what locality does it take?

\section{Can I have some feedback?}
\label{feedback}

\begin{figure}
    \centering
    \vspace{-10mm}
    \includegraphics[width=\linewidth]{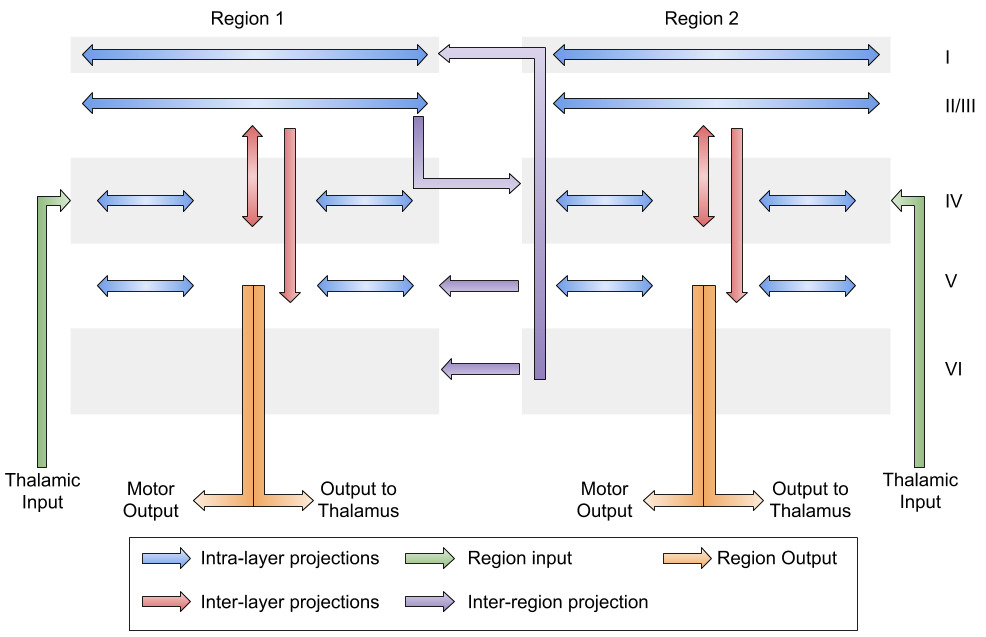}
    \caption{\footnotesize Connectivity within and between cortical regions. Each layer has a tremendous amount of intra-layer connectivity as well as projections to intra-region layers and inter-region layers. Input commonly arrives via thalamic projections and output originates in layer V targeting a variety of destinations. }
    \label{fig:layers}
\end{figure}

\lettrine[lines=3]{\initfamily\textcolor{darkgreen}{F}}{eedback} is prominent throughout the cortex. For example, in visual cortex there are ten times as many cells projecting back to the thalamus as there are feed-forward projections from the thalamus to the visual cortex \cite{sherman1986control}. A more common principle is that there are, roughly, twice as many feedback connections as feed-forward. Feedback connections are present between cortical regions as well as within these regions \cite{shipp2007structure}. Shown in Figure \ref{fig:layers} is a connectivity diagram for two neighboring regions of cortex - e.g. V1 and V2. In all areas there are exceptions, however, these connections are sufficiently regular as to generalize connectivity rules. 

The common ascending pathway starts with feed-forward thalamic input to layer IV, which then projects to layer II/III. Layer II/III will send intra-region - within region 1 - projections to layer V as well as inter-region - to region 2 - feed-forward input to layer IV. The descending pathway, starting in region 2, sends projections from its layer V and VI to region 1's layer II, V, and VI. The output from cortical regions comes via layer V. In which layer V axons bifurcate projecting to the thalamus and often a sub-cortical region associated with motor control, even in sensory regions \cite{seabrook2017architecture}. 

\begin{figure}
    \centering
    \includegraphics[width=\linewidth]{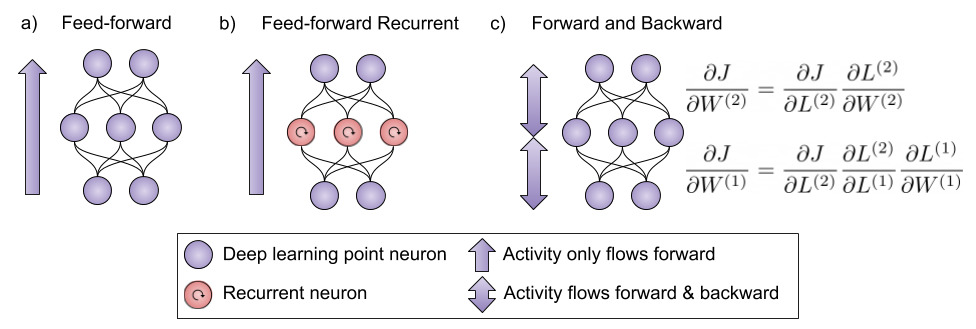}
    \caption{\footnotesize Types of connectivity in neural models. a) Standard feed-forward model. Used for non-temporal objectives - easily trained via backpropagation. b) Recurrent model. Suitable for tasks with sequential dependency - trained via backpropagation through time. c) Forward and backward model. More accurately matches cortical circuitry - onerous to train with backpropagation.}
    \label{fig:feedback}
\end{figure}

Each layer has a large number of intra-layer connections as well as forward and backward projections. Cortical layers are in stark contrast to deep learning layers, which rely on predominantly feed-forward pathways (Figure \ref{fig:feedback}a). Even recurrent models (Figure \ref{fig:feedback}b), such as the Long Short-Term Memory (LSTM) or Gated Recurrent Unit (GRU), at best approximate intra-layer connections. Deep learning models are incapable of feedback connections due to their reliance on backpropagation as a training algorithm. Differentiating the loss with respect to the parameters of a forward-backward model is shown in Figure \ref{fig:feedback}c. 

Complications arise because the activity in first layer, $L^{(1)}$, not only depends on the previous weights, $W^{(1)}$, but also on the above layers activity and weights, $L^{(2)}$ and $W^{(2)}$, respectively. Similarly, the activity in the second layer depends on its weights, and the activity and weights of layer one. This circular dependency can be problematic even for non-sequential models and implausible when applied to tasks across time.

It is important to question whether feedback provides useful capabilities or another superfluous detail. After all, solely feed-forward architectures have done remarkably well in domains like object recognition \cite{russakovsky2015imagenet}. Feedback connections are thought to provide a recirculation of activity, mechanisms for attention, and top-down expectations about incoming stimuli. Recirculating activity provides temporal awareness, as computations done in the present incorporate information from the past. Attention allows higher level regions to instruct lower level ones to selectively filter relevant aspects of incoming stimuli. Top-down expectations play an important role in developing predictive models of the world (Section \ref{sequences}). As soon as an expectation (top-down) contradicts incoming information (feed-forward), the expectation can be updated to incorporate new relevant stimuli providing better future expectations. This is not a new idea, as early as the 1940s it was proposed that vision is a result of active comparisons between sensory inputs and internal predictions \cite{eagleman2012incognito}.

In machine learning expectations are considered only at the model level. For next frame prediction, deep learning treats expectations as a distribution over pixels conditioned on the past. Expectations from the output of the model are compared with the ground truth to minimize error, globally. Similarly, deep reinforcement learning minimizes the reward prediction error when completing high level tasks. It is important to note this contradiction with biology, where expectations can occur at the network, region, layer, and even synaptic level.\footnote{This is precisely how the learning rule STDP can be used to learn temporal, predictive, sequences} 

Ignoring attention and top-down expectations, it is clear certain tasks are temporal in nature. This temporal dependency requires recurrence or information to be maintained over time, at the very least. Computational principles derived from intra- and inter-layer feedback, are likely to yield better causal models. As units of the model maintain expectations, while integrating new information to update existing beliefs. The effectiveness of attention can be increased by incorporating feedback between regions that produce attention scores\footnote{Artificial neural networks that produce soft or hard attention scores. See \cite{bahdanau2014neural} for an example} and ones that process the stimuli originally. Similarly, world models may benefit from top-down expectations - not just at the model level but within networks and layers, as well. In this way, context from higher level regions inform lower levels what is relevant to them. In turn, the lower level regions provide more meaningful information to higher regions.

\section{How does my body look?}
\label{embodiment}
\lettrine[lines=3]{\initfamily\textcolor{darkgreen}{E}}{mbodiment refers} to a hypothesized need for artificial agents to be given some type of physical presence. This manifestation is most closely associated with robotics. Where agents take the form of vehicles or humanoids, opposed to a strictly digital presence. 

The call for embodiment may be yet another anthropomorphism (Section \ref{anthro_and_approx}) of humans projecting onto artificial agents. It is not known whether embodiment is a driving factor in human cognition. However, it is clear that the brain incorporates physical movement and spatial patterns in navigation \cite{fyhn2004spatial, moser2008place}. The discovery of grid and place cells shed light on the brains ability to navigate. Allowing its host to localize itself in physical space via movement and location information. Additionally, grid cell like codes have been realized in artificial networks completing spatial navigation tasks \cite{banino2018vector}. 

It has been hypothesized that equivalent grid cells would occur in the neocortex as well \cite{hawkins2018framework}. Recent experimental evidence confirms that grid like patterns are present when navigating conceptual space \cite{constantinescu2016organizing}. In this way agents may need to derive an allocentric location signal from stimulus.\footnote{For example, in vision location information can be derived via ocular motor signals and head orientations (Figure \ref{fig:intuitive_physics}b) } So that object representations are learned in relation to one another. 

Furthermore, movement has shown to be essential in the development of vision. Kittens not allowed to move during early stages of their lives developed inadequate visual perception \cite{held1963movement}. Movement is noticeably absent from today's artificial vision systems. Convolutional neural networks, the current state of the art in computer vision, process images with a single forward pass. There are no saccades or extended visual glimpses, instead the entire image is processed uniformly to make a prediction. 

In one sense a strictly physical embodiment may not be necessary. However, pseudo embodiment, like that of conceptual space is likely required. This has the clear advantage of allowing artificial agents to store information in a navigable space. Moving through concepts in the space like one moves, physically, through a room.

\section{Anthropomorphism and Approximation}
\label{anthro_and_approx}
\lettrine[lines=3]{\initfamily\textcolor{darkgreen}{R}}{esearchers consistently describe} their artificial networks in human behavioral terms. Using phrases such as ``the neural network \textit{thinks}..." or ``the model \textit{wants}..." When in reality the network has no needs or wants, no intuition or belief. They are pattern recognition systems, operating solely on the domain which it was trained. 

While it may be easy and convenient to anthropomorphize one's models, this can be a slippery slope to which models are described under false pretenses. The danger is that by describing models with  behavioral adjectives we may overestimate their capabilities or misunderstand their aptitude. Such descriptions are likely to deceive non-technical individuals. These misunderstandings have political and economic consequences, for which researchers are responsible. Furthermore, this false confidence can have dire consequences when models are trusted with safety critical situations - autonomous vehicles or medical diagnosis.

Deep neural networks, trained via backpropagation, are in essence function approximators \cite{cybenko1989approximation}. Discretizing attributes of human cognition yields many domains which seek to approximate that specific behavior (Section \ref{descritize}). Take, for example, the case of image classification. The function we want to approximate is the \textit{behavior of vision}. Where an input stimulus (pixels) is processed and a discrete object category is identified. A neural network is trained to \textit{approximate} this behavior by minimizing a cost function over the training examples. 

However, this paradigm may be fundamentally different than that of local learning. Where learning rules depend on local temporal and spatial information. In this arrangement, behaviors arise out of network dynamics instead of via an approximation of a predefined, \textit{external}, behavior.

The approximation approach to learning must answer these critical questions: Is there a fundamental difference between approximating a behavior - via gradient descent on a cost function - and behaviors that arise out of network dynamics? Furthermore, is approximating a behavior equivalent to the behavior itself?

\section{Symbolism or Connectionism?}
\label{symbolic}
\lettrine[lines=3]{\initfamily\textcolor{darkgreen}{S}}{ymbolic or Classical AI} attempts to represent human knowledge explicitly with declarative rule based systems, logic processing, and manipulations on symbolic representations. Expert systems, fuzzy logic, knowledge bases, heuristic approaches, and graph representations are a few examples of these models. In programs like Cyc \cite{lenat1995cyc}, engineers explicitly write millions of hand programmed rules in an attempt to codify human knowledge and common sense. 

In their 1975 Turing Award Lecture, Allen Newell and Herbet Simon stated, ``symbols lie at the root of intelligent action..."\cite{newell1975computer}. Their statement is without a doubt valid, human intelligence relies on the manipulation of symbols and abstract concepts. However, the issue is with how these symbols are represented. The brain does not solve logic statements or write proofs by contradictions. Just as it does not solve search with a tree or plan with a graph. The complex cortical dynamics give rise to high level behaviors capable of the aforementioned tasks. Much like deep learning\footnote{Deep learning is, in a sense, a symbolic hybrid. As the labels required for training are symbolic, representing human semantic categories.} tries to approximate behaviors, symbolic AI abstracts and descritizes them. 

Symbolic approaches in the 60's and 70's were the norm, they are now the exception. However, deep learning suffers many of the same pitfalls of classical systems but in a more obfuscated way. For Classical AI to perform adequately, all rules or possible encounters must be codified in its knowledge base. This explicit need for predefined knowledge is paralleled in supervised learning. In this setting, we need not construct millions of if statements, instead millions of labeled training examples are required. Thus in order to build agents with common sense we would need labeled common sense data sets. This requirement is just as unfeasible as expert systems. Human knowledge can be abstracted as declarative rules but that is clearly not how the brain operates. 

Hybrid systems combine symbolic, heuristic, or classical approaches with representation learning techniques. Prominent examples of this are search engines, autonomous vehicles, and puzzle solving \cite{akkaya2019solving, silver2016mastering}. For example, AlphaGo uses deep learning to approximate value functions for actions on board configurations. A tree is constructed to unroll possible futures and search for the best estimated path in the game. 

While attempting to combine beneficial aspects from both approaches, hybrid systems are hindered by a combination of issues. First, high dimensional information, like pixels in an image, must be encoded to a symbolic representation. Second, this representation will be manipulated with predefined rules. Thus the expressivity of the system is limited by the explicit encoded rules. Third, there is the issue of feedback (Section \ref{feedback}). Symbolic representations would need to inform areas that construct representations what operations are taking place. 

Classical systems were brittle and rigid. Requirements of describing all of human knowledge became an untenable position, which is why they have mostly been abandoned. Hybrid systems offer mild robustness but in time, will fail to meet qualifications for human level intelligence. There has been ongoing debate between those who favor symbolic approaches and those who favor connectionism. This distinction is often misguided, as both paradigms can suffer similar problems. Both symbolic and connectionist approaches have relied on explicit design to solve predetermined behaviors. The real distinction must be drawn between explicit and emergent systems.

\section{What is learning?} 
\label{learning}
\lettrine[lines=3]{\initfamily\textcolor{darkgreen}{L}}{earning} is a blanket term, used to cover everything from K-Nearest Neighbor \cite{fix1951discriminatory} to the formation of new ion channels at terminal buttons of a synapse \cite{wolf2004neuroscience}. Machine learning defines it in terms of cost functions. This definition of learning operates at a macroscopic scale, modifying parameter weights in order to minimize a global cost function. Conversely, neuroscience focuses on the microscopic, defining learning in terms of ion flow, gated channels, and protein synthesis within a single neuron \cite{wolf2004neuroscience}. This section focuses on learning both at a global and local scale.

\subsection{Cost functions}
\label{cost}
In machine learning, the optimization of a cost function is synonymous with learning. Quantifying error and iteratively improving upon it, is the core of all machine learning algorithms. It has been hypothesized that the brain too implements, one or many, cost functions \cite{marblestone2016toward}. 

Most forms of learning can be framed in the concept of a cost function. For example, in biological organisms, competition between neurons enforce sparsity and specificity of neurons in response to input stimuli \cite{han2007neuronal}. This can be viewed as an optimization of its available resources to maximize information retention. Formulating a predictive model of the world can be seen as minimizing the error of future predictions \cite{sutton2018reinforcement}. In this way we can describe any learning rule or form of credit assignment as an optimization of a cost function. However, there is a difference between explicitly minimizing a cost function and framing dynamics as an optimization of a pseudo cost function.

Just as we noted in Section \ref{anthro_and_approx} that there is a fundamental difference between explicitly training a network to approximate a behavior and the behavior that arises out of network dynamics. Here we put forward that there is a similar tension between directly optimizing a cost function and network dynamics that indirectly optimize a cost function. 

\subsection{Direct vs. indirect cost functions}
\label{direct_vs_indirect}
Directly optimizing a cost function is the current approach in deep learning. Computation graphs are specified, parameters initialized, and a cost function set (e.g. mean squared error or categorical cross entropy). \textit{Learning} happens via backpropagation and some flavor of stochastic gradient descent \cite{kingma2014adam, zeiler2012adadelta, qian1999momentum}. In which, the gradients of the cost function are computed with respect to each parameter in the graph. The parameters are updated as training batches are seen by the network. Gradient descent works by updating parameters in a way such that it minimizes the cost function for all examples in the training data. 

Indirect optimization involves constructing network dynamics in such a way (i.e., local learning rules \cite{hebb1949organization, gerstner1996neuronal, markram1995action}) that they approximate the minimization of some cost function, without being explicitly set to do so. For example, STDP can be interpreted as approximate gradient descent on an objective function \cite{bengio2015towards}. In this way, STDP does not \textit{explicitly} minimize an objective function, however, its dynamics may be described in such a way to frame it as an optimization on an arbitrary cost function. 

At the current point in time, is unclear which of these two paths is more fruitful and if there is even a difference between them. Neuroscience should investigate dynamics of cortical networks, to answer if the brain does in fact optimize cost functions \cite{marblestone2016toward}, and if so which kind. Research in artificial networks should continue to investigate indirect cost functions as to avoid narrow, domain specific, solutions.

\subsection{Predictive sequences}
\label{sequences}

Learning predictive sequences is a well established trait of intelligence \cite{sutton1988learning, rao2000predictive, sutton2018reinforcement, lake2017building}. In the field of reinforcement learning this manifests itself in temporal difference learning \cite{sutton1988learning}. In biology this trait has obvious evolutionary advantages. If an organism can predict future events it is more likely to survive than ones that cannot. 

This property of human intelligence can be observed introspectively and in juxtaposition with current artificial agents. Analyzing ones morning routine, tasks such as pouring coffee are completed with minimal cognitive resources. Conversely, an artificial agent (i.e. robot) performing this task does so via a discrete number of segmented objectives\footnote{In hierarchical reinforcement learning this may be viewed as one global task composed of smaller ones}: \texttt{move gripper from position $(x_1, y_1, z_1) \rightarrow (x_2, y_2, z_2)$, rotate gripper and grasp handle, lift pot to position $(x_3, y_3, z_3)$, visually locate target mug, etc.}. No matter how many times the artificial agent completes these tasks, the difficulty will not decrease, the cognitive computations required do not decrease, and the smoothness of actions remain rigid and discontinuous. 

However, we know from observational and experimental trials the opposite are all true for humans. The more times you practice motor tasks, the easier and less rigid the motion becomes. Additionally, humans use less cognitive resources for learned tasks opposed to novel and difficult ones \cite{van1998changes, petersen1998effects, sakai1998transition, eagleman2015brain}. 

Zooming in to the microscopic level we can see how this could be implemented computationally. Evidence suggests STDP is well suited to learn temporal sequences \cite{rao200216}. In order for a pre-synaptic weight to be strengthened it must be active prior to the post-synaptic cell firing \cite{markram1997regulation}. This time sensitive nature of synaptic learning inherently gives it the ability to learn temporal sequences \cite{rao200216, minai1993sequence, rao2000predictive}. In this way pre-synaptic weights learn to predict which post-synaptic cells will become active. Many neurons in a cortical network will then learn a predictive sequence, each one predicting the next. This property is further elaborated in Section \ref{intuitive} and Figure \ref{fig:intuitive_physics}.

Zooming back out we can now see how our coffee example is satisfied. The more attempts at this motor task, the better synaptic weights become at predicting the next active neuron. Therefore, less resources are used to predict erroneous neurons and electrical activity is more streamlined through correct neurons which increases the ease and smoothness of the task.

Connecting ideas from Section \ref{direct_vs_indirect}, STDP works locally (i.e. at the synapse), where its dynamics yield sequential modeling between the pre- and post-synaptic cell. Globally, this manifests an indirect optimization of entropy within neuronal populations and yields predictive sequence learning as a functional result. 

\subsection{Sensory motor integration}
\label{smi}

An animal's interaction with the environment takes place through two primary mechanisms. Organisms receive input via their sensory receptors. These signals provide the organism with information about the environment at the current time. The output from an organism allows it to move and manipulate its environment. Motor commands result in changing limb and body orientation in ways that may deform the environment. Motor commands effect where sensory signals will be captured and sensory signals, in turn, inform where to move those sensors. 

This relationship is best understood in an example of vision. As one reads a line of text, the eyes receive visual information - a sensory signal. This stimuli is processed and the eyes are instructed to continue moving - a motor signal. Each time, a motor command is conditioned on previous sensory information. So that eyes move across the line to capture the next word in the sentence.

In the cortex, the integration of sensory and motor signals is well documented. The common consensus is that sensory regions and motor regions are distinct entities. In order to activate movements, neurons in motor cortex must be activated. However, recent evidence suggests the distinction is less clear. Activity in sensory regions has been found to strongly correlate with limb movement \cite{karadimas2019sensory}. These observations are supported by bifurcating axons from sensory regions (Figure \ref{fig:layers}). For example, the axons descending from layer five in the visual cortex, split and send projections to the lateral geniculate nucleus and superior colliculus \cite{seabrook2017architecture}. The latter of which is associated with head and eye movements. As a result, even low level visual regions, like V1, have influence over saccades.

This suggests that sensory and motor commands are more intertwined than previously thought. This is further evidence that we need not separate these tasks (Section \ref{descritize}). Instead, sensory and motor commands are codependent, often occurring within the same cortical region, and possibly the same layer. The synergistic interaction of sensory and motor aspects allows each to have information about the other. So that when motor commands are issued to move the eyes, there is a sensory expectation of what will be seen. Conversely, deep reinforcement learning maintains a policy,\footnote{A policy specifies a distribution over actions conditioned on an observation, denoted $\pi(a|o)$. Taking actions is equivalent to issuing motor commands} that takes sensory input and produces motor actions. However, the information only flows one way (Section \ref{feedback}). Sensory regions, in the policy network don't receive feedback. Therefore, they have no information regarding which action was taken and, as a result, no expectation of what sensory information will come next.

\section{How can we be more data efficient?}
\label{data}
\lettrine[lines=3]{\initfamily\textcolor{darkgreen}{C}}{urrent deep learning} models are highly inefficient when learning from data. They require hundreds of thousands of training examples to learn object categories. Where as humans can learn after a single example. In reinforcement learning, the situation is orders of magnitude worse. Often artificial agents require millions of attempts, in simulated environments, to master trivial tasks. And in most cases knowledge learned in one environment does not transfer to others. Additionally, learning new environments often decreases performance in previously learned ones (catastrophic forgetting). 

\subsection{Causes of inefficiency}
Slow learning speeds may be a result of the training paradigm and the representation style of current deep networks. The gradient descent algorithm must make infinitesimal changes to weights because the gradient is only known locally. Large changes result in poor performance and may cause an explosion in weight values. Slowly, the network must iteratively descend the cost function in parameter space. Each example updates the weights with a small change. As a result a deep network never learns instantaneously. Additionally, this process only provides support for examples found in the training data. Consequences of leaving the training manifold were presented in Section \ref{adversarial}. This, slow, iterative process of gradient descent contributes to data inefficiency by requiring more examples to update weight values. 

In addition to gradient descent, dense codes contribute to the issues listed above. Current neural networks represent weights and activities as real, floating point numbers. As a result these high dimensional representations must be slowly pulled apart, via gradient descent, to obtain separation between categories (e.g. different object categories in image recognition). As a result they are highly susceptible to noise and fail to provide support for all spaces in the input domain. Intuitively this can be realized as a result of their near infinite capacity. It is difficult to learn smooth boundaries in these high dimensional spaces. 

Conversely, as mentioned in Section \ref{encode}, sparse codes sacrifice an infinite capacity for robustness and speed \cite{foldiak2003sparse}. Ideas from learning predictive sequences (Section \ref{sequences}) can be combined with top-down expectations (Section \ref{feedback}) to learn predictive models of the environment. In this way we can combine the beneficial properties of sparse representations with predictive models. This combination will yield faster learning and a model of the environment. Similar to model based reinforcement learning, models help us to predict, plan, and take actions all without using more costly real world samples - increasing data efficiency. 

\subsection{Priors}

The mammalian brain comes preloaded with a variety of features at birth. These attributes are not learned but are innate to the animal. For example, cells in the cat visual cortex have fully formed receptive fields immediately after birth \cite{hubel1963receptive}. These cell's response properties closely match those of adult cats. Similarly, human infants are predisposed to recognize faces \cite{mondloch1999face}. This would suggest that learning visual fields and low level stimuli are not necessary. Instead, the problem becomes how to interpret these high level representations and predict their occurrence.

To increase data efficiency we can impose knowledge on models in the form of priors. A prior many be an anatomical or physiological trait innately built into the model. For example, two dimensional local connectivity is better suited for vision than one dimensional full connectivity. Visual stimuli, such as images, have local and spatial relationships between pixels. This type of prior led to the creation of the first convolutional models \cite{fukushima1988neocognitron, lecun1998gradient}. In this manner, we may extract other computational principles from biology and impose them as priors in artificial models. 

The features discussed in Section \ref{robust_biology} can all be interpreted as priors. Specifically, inhibition leads to the construction of receptive fields. The same receptive field structure is found in vision as well as somatosensory regions. Neighboring areas in cortex share similar receptive fields \cite{mountcastle1957modality}. Stimuli centered in one receptive field will cause inhibition in a neighboring field \cite{wolf2004neuroscience}. This process serves to sharpen the response properties of stimuli in corresponding receptive fields.

Deep learning's approach would be to train individual models - one for faces, another for somatosensory, a third for intuitive physics, etc. If models are trained for specific tasks they will only be capable of those tasks. The idea of imposing priors is more general. It predisposes certain regions of a cortical network to be well suited for a specific, without restricting them to that task. The same area that handles problem solving also tackles planning, social behavior, and emotion. Creating explicit models for each faculty will likely miss important intricacies of how these abstract concepts interweave with one another. 

Deep learning may be able to engineer more efficient flavors of gradient descent \cite{qian1999momentum, kingma2014adam}. However, it is likely constrained by dense codes, small gradient updates, and the requirement to learn \textit{optimal} deep targets. Again, while backpropagation has proven immensely useful for optimization, the algorithm constricts the creation of certain classes of models while requiring unreasonable amounts of training data. Priors can be induced on models in the form of structural or functional specifications. These attributes make learning all representation levels easier and, in some cases, unnecessary. As a result we can learn faster, develop predictive models, and ultimately be more data efficient. 

\section{What's common sense?}
\label{common_sense}

\lettrine[lines=3]{\initfamily\textcolor{darkgreen}{D}}{eep learning models} offer the appearance of intelligence without understanding. Numerous examples can be found in object recognition and image captioning \cite{lake2017building}. It quickly becomes clear that these models are in fact not learning physical objects or how to describe them. Many times researchers have called for \textit{common sense} to be incorporated into models \cite{marcus2018deep, lake2017building}. 

A high level, psychological, treatment of ``ingredients for building more human-like learning" was presented in \cite{lake2017building}. The authors recirculated ideas of endowing models with ``causality", "intuitive physics", and ``intuitive psychology" but provided no insight into how this could be accomplished. Here we will layout an example of intuitive physics derived via sparse codes (Section \ref{encode}), predictive sequences (Section \ref{sequences}), and sensory motor integration (Section \ref{smi}). Following this we will tie in concepts from previous sections and contrast other approaches. 

\subsection{Intuitive Physics}
\label{intuitive}

\begin{figure}
    \centering
    \includegraphics[width=\linewidth]{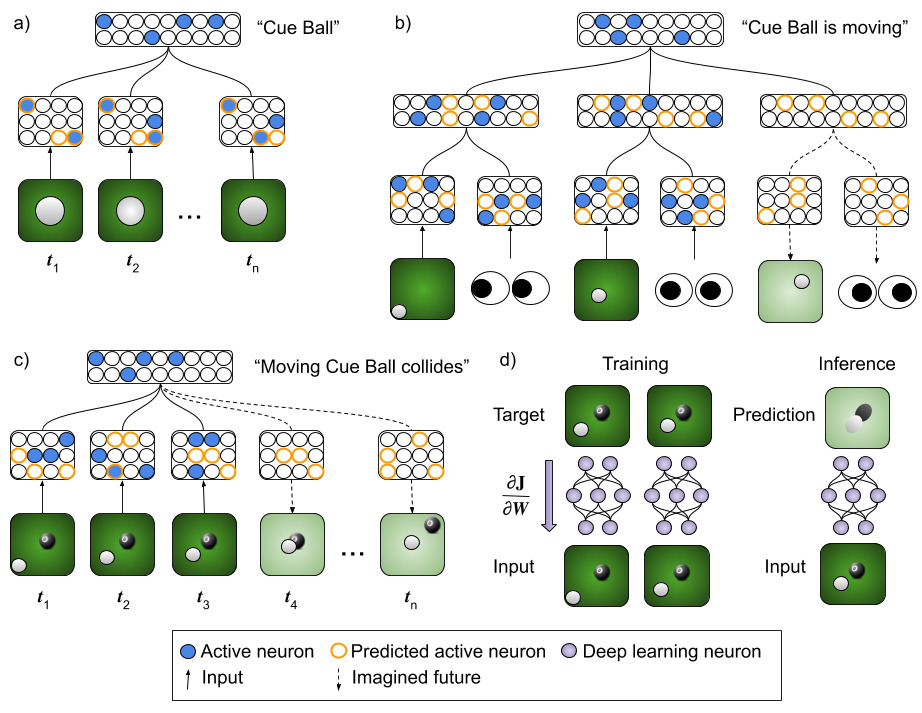}
    \caption{\footnotesize How to capture intuitive physics with predictive sequence memory. a) Several glimpses of the same object will activate different neurons. This sequence of varying activity forms a stable representation that we interpret as the cue ball. b) Visual and motor input now form the input sequence. Each time we have an expectation of what we will see conditioned on the location of the eyes. The motor and visual sequence informs us of the high level representation that the ball is moving. c) Incorporating movement information from part b the system predicts next active neurons corresponding to a collision between balls. d) The deep learning approach. The training phase would consist of minimizing the mean squared error of predicted pixel intensities with the ground truth label. At inference time we show the current frame and the network predicts the next frame.}
    \label{fig:intuitive_physics}
\end{figure}

Predictive sequence learning, as described in Section \ref{sequences}, gives agents the ability to predict at a microscopic scale - learning to predict the next active neuron. Learning causal models or intuitive physics involves predicting at a macroscopic scale - the interaction between complete physical objects. This type of inference requires learning hierarchies of sequences - sequences of sequences. 

Let's examine what is required to know the result of a collision in billiards, when the cue ball hits the eight ball. In order to learn the ``intuitive physics" of this interaction a model first needs the ability to form predictive sequences (STDP from Section \ref{sequences}) and incorporate sensory signals to produce motor commands (Section \ref{smi}). Vision consists of constantly moving the eyes and incorporating these glimpses over time. Figure \ref{fig:intuitive_physics} displays the following example of intuitive physics. Neurons in these diagrams may get input from feed forward basal dendrites, intra layer connections, or apical dendrites (common sources of feedback from higher layers). The exact wiring is not relevant for the example,\footnote{This should not be interpreted as a literal architecture. The details are far more complex} instead the focus is on sparse vectors encoding information and active neurons predicting the next neuron to become active.

(1) Identifying the cue ball at rest is the first temporal sequence we need. As one looks at the cue ball light patterns consistently hit differing cells in the retina causing variable signals to the visual cortex over time. This is shown in Figure \ref{fig:intuitive_physics}a where glimpses of the cue ball activate different neurons early in the hierarchy. Despite these constantly changing signals, one perceives stability and knows the identity of the cue ball (high level representation in Figure \ref{fig:intuitive_physics}a). This provides us with a many-to-one mapping: where a sequence of visual stimuli map to one object representation, the cue ball.

(2) As the cue ball begins moving we invoke another level in the sequence hierarchy. Not only do we have a sequence of stimuli about the cue ball but we also have a sequence of changing location information. Location information can be derived from ocular motor signals as the eyes track the cue ball through space (Figure \ref{fig:intuitive_physics}b). The combination of visual and location information is combined to form a new sequence, informing us the cue ball is in motion. As neurons have already predicted the next active neuron, the contextual location information allows them to predict not only the cue ball but also the next location.

(3) The final step is the incorporation of the eight ball (Figure \ref{fig:intuitive_physics}c). Unrolling our prediction about the cue ball in conjunction with location information (part 2), we know the trajectory of the ball. This process predicts the location of the cue ball at a simultaneous location of the 8 ball. From past experiences of objects in motion reaching simultaneous locations we have a prediction (i.e. future active neurons are predicted) of the consequence of this sequence. 

In Figure \ref{fig:intuitive_physics}b and \ref{fig:intuitive_physics}c imagined futures correspond to what the agent thinks will occur next. This prediction is \textbf{not} a direct reconstruction like Figure \ref{fig:intuitive_physics}d. In deep learning one would explicitly train a network to predict next frames in a sequence \cite{srivastava2015unsupervised}. However, this would clearly miss important aspects of the temporal hierarchy described above. Frames predicted by the model would most likely appear blurry as the network cannot grasp abstract concepts of objects, locations, and their interaction.

\section{How to remember?}
\label{memory}
\lettrine[lines=3]{\initfamily\textcolor{darkgreen}{M}}{emory} is an essential aspect of human cognition. Memory is not exact, instead it is a subjective, compressed, representation of a stimulus or sequence of stimuli.\footnote{Memories can invoke the same temporal hierarchy discussed in Section \ref{intuitive}. Where a single thought may recall a whole sequence of past events.}

The human brain implements an associative reference memory. Where thoughts in the present can recall related memories of the past. The brain accomplishes this in constant time, $O(1)$. Where many keys - neural activity - can map to single values (e.g. memories in hippocampus). This efficient routing mechanism is in stark contrast to current memory systems for neural networks \cite{graves2014neural}. For example, in a Neural Turing Machine (NTM) accessing memory is an $O(n)$ operation, where a context vector - current activity - is compared to all previously stored memories. Not only is this process highly inefficient but it can dilute the content of the important memory to be recalled. Dilution can occur as a result of the weighted sum implemented by soft attention. Conversely, sampling a single item, hard attention, is not differentiable resulting in less frequent use in practice.

There is also the issue of the temporal scale at which memory takes place. Synaptic modifications, in the CA region of the hippocampus, are well documented. Processes like long term potentiation (LTP) and depression (LTD) are known to strengthen synapses over longer periods of time \cite{wolf2004neuroscience}. Leading to the formation of more long term memory. However, the mechanisms by which LTP occurs can be fast acting on the scale of seconds \cite{wolf2004neuroscience}. These more rapid changes can produce short term memory. In this way, models have the ability to operate memory at different timescales. 

Current deep learning approaches rely on distinct training and testing phases. During training, a synchronous update is performed across the entire network. Once training is completed, weights are fixed and no modifications occur. This paradigm poses two issues. First, networks will only learn at a single temporal scale. Hierarchical models attempt to remedy this by constructing explicit models (Section \ref{descritize}) for different temporal scales. For example, one model handles long term themes while the other focuses on small intervals that constantly change  \cite{kulkarni2016hierarchical}. Second, if inputs during the testing phase deviate from the training data the network will have no ability to adapt. This problem is amplified in sequential modeling, where information at one time step should affect how the information at the next step is processed. This has led to the creation of networks with modifiable weights \cite{ba2016using}. 

Memory is an integral part of cognition. It allows for information to be compressed, stored, and recalled when stimuli in the present associate with its content. The temporal scale and computational requirements are important aspects of artificial memory construction - humans could not function with an $O(n)$ memory. 

\section{Conclusion}
\lettrine[lines=3]{\initfamily\textcolor{darkgreen}{M}}{any} of the questions addressed in this article are intertwined with one another. The externality of rewards and goals (Section \ref{rewards}) is associated with a direct cost function (Section \ref{direct_vs_indirect}). Sparse coding schemes (Section \ref{encode}) can aid in building robust models (Section \ref{robust}). Discretizing intelligence (Section \ref{descritize}) leads to the approximation of intelligent behaviors (Section \ref{anthro_and_approx}) which is accomplished by direct cost functions (Section \ref{direct_vs_indirect}). The formalism of learning rules (Section \ref{credit_assignment}) is cleary tied with learning (Section \ref{learning}); and deep targets (Section \ref{deep_targets}) imply approximation (Section \ref{anthro_and_approx}) via a direct cost function (Section \ref{direct_vs_indirect}). Feedback pathways (Section \ref{feedback}) are involved in sensory motor integration (Section \ref{smi}) which plays a role in allocentric location signals for conceptual space (Section \ref{embodiment}). Predictive sequence learning (Section \ref{sequences}) is enveloped in sensory motor integration (Section \ref{smi}) which can be used for understanding causal models (Section \ref{intuitive}) and solving many forms of invariance (Section \ref{invariance}). Greater data efficiency (Section \ref{data}) can be achieved by alternative coding schemes (Section \ref{encode}), memory utilization (Section \ref{memory}), and learning predictive models of the world (Section \ref{intuitive}).

Researchers can list off properties of human cognition which the machine learning community will eagerly solve. However, this leaves us to question what we're building via this approach. Are we creating agents that understand causality and the consequences of interactions, or are we creating function approximators on a data manifold?

There are numerous issues AI researchers must address when constructing intelligent agents. Often those in the machine learning community shy away from neuroscience. The endless complexity is quite daunting and it is never clear which features are essential and which are superfluous. However, capturing computational principles from the brain may prove essential in our future work. 

Regardless, machine learning approaches need to think critically about a number of issues. It is imperative to resist the urge to anthropomorphize our models. Not only does this misrepresent our models capabilities but may deceive others. We need to consider what type of learning we're trying to accomplish and what mathematical form that should take. Cost functions are useful in order to quantify attributes of the learning process. However, directly optimizing them will produce rigid and narrow models. Sequences, feedback, and sparsity - all of which are important aspects in human cognition - have been relatively ignored due to complexities with backpropagation through time and non differentiability. Backpropagation is placing constraints on the types of models we are able to build which limits progress. It is important to consider the locality of models. Rewards, goals, and data manipulations are all external in machine learning. Conversely, in biology these process occur within the organism. 

These topics provide avenues to integrate principles from neuroscience into machine learning. It is unlikely that every minute detail - down to the protein, molecule, and ion - will need to be implemented to create intelligence. However, there are clear computational principles on which the brain operates. The problem is finding these computational needles in a haystack of biological complexity. Biology has clear constraints but by not using it as a guide we are constraining ourselves. 

\printbibliography

\end{document}